\documentclass[10pt,letterpaper]{article}

\usepackage{graphicx, amsmath, amssymb, caption, multirow, overpic, textpos}
\usepackage{color}
\usepackage{array}
\usepackage[pagebackref=false, breaklinks=true, colorlinks=true, bookmarks=false, urlcolor=blue]{hyperref}

\usepackage{ccn}
\usepackage{pslatex}
\usepackage{cite}

\newlength\savewidth\newcommand\shline{\noalign{\global\savewidth\arrayrulewidth
  \global\arrayrulewidth 1pt}\hline\noalign{\global\arrayrulewidth\savewidth}}
\renewcommand{\paragraph}[1]{\vspace{1.25mm}\noindent\textbf{#1}}
\newcommand{\tablestyle}[2]{\setlength{\tabcolsep}{#1}\renewcommand{\arraystretch}{#2}\centering\footnotesize}

\newcolumntype{x}[1]{>{\centering\arraybackslash}p{#1pt}}
\newcolumntype{y}[1]{>{\raggedright\arraybackslash}p{#1pt}}
\newcolumntype{z}[1]{>{\raggedleft\arraybackslash}p{#1pt}}

\newcommand{\authorskip}{\hspace{2.5mm}}

\title{A Parameter-efficient Multi-subject Model for Predicting fMRI Activity}
 
\author{
    {\large \bf Connor Lane (connor.lane@childmind.org)} \authorskip {\large \bf Gregory Kiar (gregory.kiar@childmind.org)} \\
    Center for Data Analytics, Innovation, and Rigor (DAIR) \\
    Child Mind Institute
}

\begin{document}

\maketitle

\section{Abstract}
{
\bf
This is the Algonauts 2023 submission report for team ``BlobGPT''.  Our model consists of a multi-subject linear encoding head attached to a pretrained trunk model. The multi-subject head consists of three components: (1) a shared multi-layer feature projection, (2) shared plus subject-specific low-dimension linear transformations, and (3) a shared PCA fMRI embedding. In this report, we explain these components in more detail and present some experimental results. Our code is available at \url{https://github.com/cmi-dair/algonauts23}.
}

\section{Motivation}

The standard approach for predicting fMRI activity from image stimuli is to first extract features from a pretrained deep neural network, and then fit a separate linear regression model for each subject to map from feature space to subject fMRI activity space \cite{naselaris2011encoding}. One challenge with this approach is that both the deep net features and fMRI activity vectors have 10K+ dimensions. A naive linear encoding model would require 100M+ parameters per subject. Our main motivation was to build a more parameter-efficient linear encoding model. We achieved this through (structured) dimensionality reduction, and by sharing the components of our model across subjects where possible.

\section{Model architecture}

\begin{figure}[t]
\centering
\includegraphics[width=0.45\textwidth]{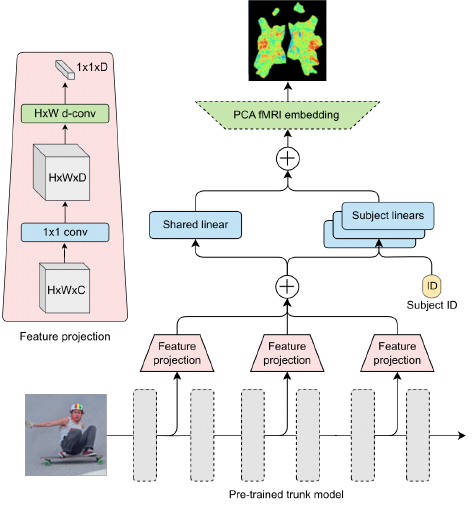}
\caption{\textbf{Model architecture.} Overview of the multi-subject linear fMRI encoding head. Intermediate feature tensors are extracted from a pretrained ``trunk'' model, projected to a lower dimensional latent space, and averaged. Next we apply shared and subject-specific linear transformations and add the results. Last, we reconstruct the full fMRI activity vector using a frozen group PCA embedding. The feature projection module separately projects along the channel dimension (1$\times$1 convolution) then pools over the height and width (depth-wise convolution with kernel size = stride = H$\times$W). Dashed lines indicate modules frozen by default.} 
\label{fig:arch}
\end{figure}

Our model consists of a multi-subject linear fMRI encoding ``head'' that is attached to a pretrained ``trunk'' model (Figure~\ref{fig:arch}). The input(s) to the encoding head are full-size feature tensors (height $\times$ width $\times$ channels) from one or more intermediate layers of the trunk model. The output of the encoding head is a predicted fMRI activity vector.

\paragraph{Feature projection.}
The first stage of the encoding head linearly projects the high-dimensional feature tensors down to a manageable dimension (e.g.\ 1024). Each input layer is projected independently and the resulting latent vectors are batch-normalized and then averaged. As a result, each latent dimension aggregates information from multiple layers. 

To conserve parameters, the feature projection module first projects to the target latent dimension along the channels axis using a 1$\times$1 convolution, then pools over the height and width axes using a separate learned pooling map for each latent dimension (Figure~\ref{fig:arch} inset; see also \cite{st2023brain}). Equivalently, the spatial pooling can be viewed as depth-wise convolution \cite{chollet2017xception} with kernel size = stride = height $\times$ width. Yet another way to view this transformation is as a standard linear projection where the projection weight for each latent dimension is factorized as the product of a spatial map (height $\times$ width $\times$ 1) and a channel filter (1 $\times$ 1 $\times$ channels). This efficient structured projection provides roughly a 1000x parameter savings compared to standard linear projection, e.g.\ PCA.

Importantly, the feature projections are learned jointly and shared across subjects. This reflects the hypothesis that there should be a common visual feature space that universally captures fMRI activity across people.

\begin{figure}[t]
\centering
\includegraphics{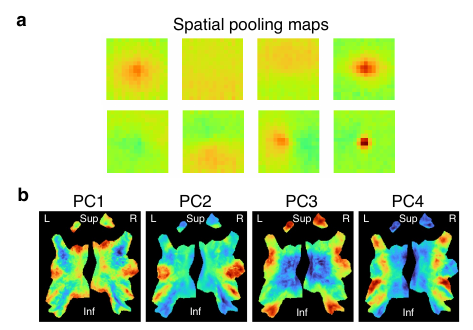}
\caption{\textbf{Spatial pooling and principal component maps.} \textbf{a} Exemplar spatial pooling maps computed by Gaussian mixture clustering. \textbf{b} Group principal component maps.} 
\label{fig:maps}
\end{figure}

In Figure~\ref{fig:maps} \textbf{a}, we visualize several exemplar spatial pooling maps found by Gaussian mixture clustering. Interestingly, the exemplar maps show selectivity for different spatial scales as well as upper/lower and left/right hemi-fields.

\paragraph{Low-dimensional encoding.}
The middle stage of our model maps from the low-dimensional latent feature space to a low-dimensional latent activity space. It consists of both shared and subject-specific linear mappings, with the results of each added together. The subject ID is used to select among the subject-specific mappings. In addition, it is possible to predict subject-agnostic ``group average'' activity patterns by bypassing the subject-specific pathway altogether.

This is the only component of our model that learns a specialized transformation for each subject. Crucially, the number of subject-specific parameters is typically very small (e.g.\ 2M). As a result, the model can in principle be adapted for new subjects relatively easily.

\paragraph{PCA fMRI activity embedding.}
The last stage of the encoding head reconstructs the full 40K-dimensional fMRI activity vector from the low-dimensional activity latent using an affine embedding. Due to the high dimensionality of the fMRI activity, this embedding contains the majority of our model's parameters. To reduce the risk of overfitting, we initialized the embedding using PCA and froze it during training. Specifically, we performed PCA on the combined fMRI activity data from all 8 subjects. We then set the embedding weight and bias to be the PCA basis and center respectively.

Figure~\ref{fig:maps} \textbf{b} shows the first 4 principal components, projected onto a cortical flat map \cite{gao2015pycortex}. The maps for these components are very smooth with high loading on the most robustly activated visual regions (face, body, and place selective regions; early visual cortex).

\paragraph{Parameter count.}
The total parameter count for the base configuration of our multi-subject encoding head (Table \ref{tab:arch_params}) is 106M, with 25M trainable parameters (excluding the frozen PCA embedding). A naive linear regression model with the same input features would require roughly 47B parameters.

\begin{table}[t]
\tablestyle{6pt}{1.02}
\scriptsize
\begin{tabular}{y{48}|y{116}}
config & value \\
\shline
trunk & \verb|eva02_base_patch14_224.mim_in22k| \\
layers & \verb|blocks.{0,2,4,6,8,10}| \\
latent dimension & 1024 \\
PCA dimension & 2048 \\
\end{tabular}
\vspace{-.5em}
\caption{\textbf{Architecture parameters.}
\label{tab:arch_params}}
\end{table}

\begin{table}[t]
\tablestyle{6pt}{1.02}
\scriptsize
\begin{tabular}{y{96}|y{68}}
config & value \\
\shline
optimizer & AdamW \\
batch size & 512 \\
learning rate & 6e-4 \\
beta1 & 0.9 \\
beta2 & 0.99 \\
weight decay & 0.8 \\
feature dropout & 0.9 \\
crop scale & 0.8 \\
training steps & 5000 \\
warmup steps & 250 \\
learning rate schedule & cosine decay \\
min learning rate & 3e-5 \\
\end{tabular}
\vspace{-.5em}
\caption{\textbf{First phase.} Linear encoding head only.
\label{tab:first_phase}}
\end{table}

\begin{table}[!h]
\tablestyle{6pt}{1.02}
\scriptsize
\begin{tabular}{y{96}|y{68}}
config & value \\
\shline
optimizer & AdamW \\
batch size & 192 \\
learning rate & 1e-5 \\
beta1 & 0.9 \\
beta2 & 0.99 \\
weight decay & 0.8 \\
feature dropout & 0.9 \\
crop scale & 0.8 \\
training steps & 2000 \\
warmup steps & 100 \\
learning rate schedule & cosine decay \\
min learning rate & 0.0 \\
\end{tabular}
\vspace{-.5em}
\caption{\textbf{Second phase.} End-to-end fine-tuning.
\label{tab:second_phase}}
\end{table}

\section{Implementation details}

\paragraph{Trunk model.}
For our best submission model, we used an EVA02 vision transformer as our pretrained trunk model \cite{fang2023eva02}. Specifically, we used the \texttt{eva02\_base\_patch14\_224.mim\_in22k} model available in Pytorch image models (timm) \cite{wightman2019pytorch} (image size=224, patch size=14, embed dim=768, depth=12, params=86M). The EVA02 family of models are pretrained with masked image modeling \cite{bao2021beit} using a large EVA-CLIP encoder  \cite{fang2023eva} as a teacher model. Importantly, since CLIP models are trained to align representations across images and associated captions \cite{radford2021learning}, this means that the intermediate features in EVA02 are implicitly shaped by linguistic supervision.

\paragraph{Data preparation.}
We trained our models using only the Algonauts 2023 Challenge data \cite{gifford2023algonauts}. For each subject, we randomly split the provided ``training'' set into private train (85\%), validation (10\%) and test (5\%) subsets, and then combined these across subjects. For training, images were random resize-cropped to the model input size (224 $\times$ 224) using a large crop ratio (e.g.\ 0.8) and a square aspect. Intuitively, this simulates subjects' slight variation in fixation point. Other data augmentations provided no benefit and in some cases hurt performance (e.g.\ horizontal flip). For evaluation, images were only resized. The fMRI activity targets for each subject were mapped to a common region of interest (ROI) space defined as the union of all individual subject challenge ROI masks. Missing vertices for individual subjects were filled with zeros.

In parallel, we also organized a compressed version of the full NSD dataset \cite{Allen2022}, including downsampled images, flat-map projected fMRI activity maps \cite{gao2015pycortex}, and COCO image annotations. The dataset is publicly available on the Huggingface datasets repository (\href{https://huggingface.co/datasets/clane9/NSD-Flat}{clane9/NSD-Flat}).

\begin{figure}
\centering
\includegraphics{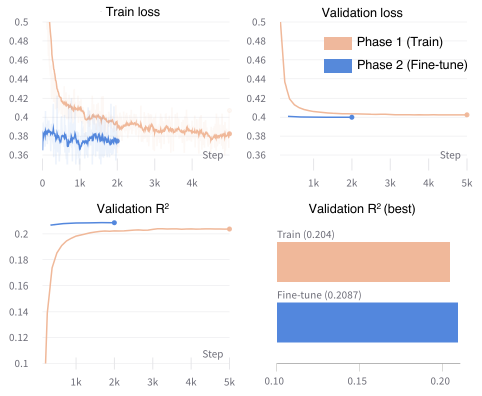}
\caption{\textbf{Loss curves for phase 1 (train) and phase 2 (fine-tune).} The metrics shown are mean squared error (MSE) loss on the train and validation splits, and the median $R^2$ on the validation split. Note that Phase 2 is initialized with the best checkpoint from Phase 1.}
\label{fig:train_curves}
\end{figure}

\paragraph{Training recipe.}
We trained our model to reconstruct fMRI activity using the standard mean squared error (MSE) loss. We regularized our model using weight decay and dropout applied to the features extracted from the trunk model.

We trained our model in two phases. First, we trained the multi-subject encoding head from random initialization with the trunk model and final fMRI PCA embedding frozen (Table~\ref{tab:first_phase}). Next, we unfroze the self-attention blocks of the EVA02 trunk and fine-tuned the model end-to-end \cite{touvron2022three} (Table~\ref{tab:second_phase}).

We tuned the hyperparameters of our model for the first phase using roughly 50 trials of Bayesian optimization \cite{falkner2018bohb} implemented in WandB \cite{wandb}. We found that having sufficient regularization was key for good performance. As a result, our model was trained with relatively high amounts of weight decay and feature dropout. We did no hyperparameter tuning for the second fine-tuning phase. Instead, we adapted the fine-tuning recipe from \cite{cherti2023reproducible}.

We trained our model using a single NVIDIA V100 32GB GPU. We accelerated model training by using automatic mixed precision and by loading all data into system memory up front. Training wall time was roughly 2 hours for the first phase and 1 hour for the second phase.

\section{Experimental results}

\paragraph{Overall performance.}
Figure~\ref{fig:train_curves} shows the overall performance of our model. Our final median reconstruction $R^2$ (over subjects and vertices; un-normalized) was 0.204 after Phase 1 and 0.2087 after Phase 2. The corresponding challenge scores (mean noise-normalized $R^2$) were 0.589 and 0.602.

Interestingly, we observed only a very modest 2\% improvement from the Phase 2 end-to-end fine-tuning. In further experiments, we also observed a tendency to overfit when fine-tuning more trunk parameters and/or for a longer schedule.

\paragraph{Trunk comparison.}
In Table~\ref{tab:trunk_comp}, we compare the performance of our multi-subject encoding head trained with different trunk models. The trunks all use the ViT-base architecture \cite{dosovitskiy2020image}, but vary on pretraining task and dataset. Consistent with recent work \cite{conwell2022can}, we observe a trend for trunks trained with larger and more diverse datasets to perform better. For example, the two trunks trained only on ImageNet-1K perform the worst overall.

However, there also appears to be an effect of pretraining task. For example, the EVA02 model (our base trunk) is trained only on ImageNet-22K, which contains 14M images. Yet it outperforms models trained on carefully curated datasets 10-100x as large.

Importantly, EVA02 does implicitly benefit from larger data. Its teacher model was pretrained on a combination of image datasets totaling 30M images, and then fine-tuned with a CLIP objective on LAION-400M \cite{fang2023eva,schuhmann2021laion}. Nonetheless, this hints that a rich top-down supervision signal may be just as important for learning brain-like representations as a large and diverse input dataset.

\begin{table}[t]
\tablestyle{6pt}{1.02}
\begin{tabular}{y{80}y{80}x{30}}
method & dataset & $R^2$ \\
\shline
MAE \cite{he2022masked} & IN-1K & 0.1832 \\
DINO \cite{caron2021emerging} & IN-1K & 0.1863 \\
Supervised \cite{wightman2019pytorch} & IN-21K+IN-1K & 0.1922 \\
CLIP \cite{radford2021learning} & CLIP-400M & 0.1958 \\
DINOv2 \cite{oquab2023dinov2} & LVD-142M & 0.1968 \\
CLIP \cite{radford2021learning} & LAION-2B & 0.1971 \\
BEiTv2+Sup. \cite{peng2022beit} & IN-1K+IN-22K & 0.1984 \\
EVA02 \cite{fang2023eva02} & IN-22K & 0.2019 \\
EVA02+CLIP \cite{fang2023eva02} & IN-22K+Merged-2B & 0.2023 \\
\end{tabular}
\vspace{-.5em}
\caption{\textbf{Trunk comparison.} Median $R^2$ for various trunks trained using different pretrain tasks and datasets. All trunks use the ViT-base architecture \cite{dosovitskiy2020image} and are available in timm \cite{wightman2019pytorch}. Phase 1 training only (no trunk fine-tuning) using a shortened schedule (1250 steps). \label{tab:trunk_comp}}
\end{table}

\begin{figure*}
\centering
\includegraphics{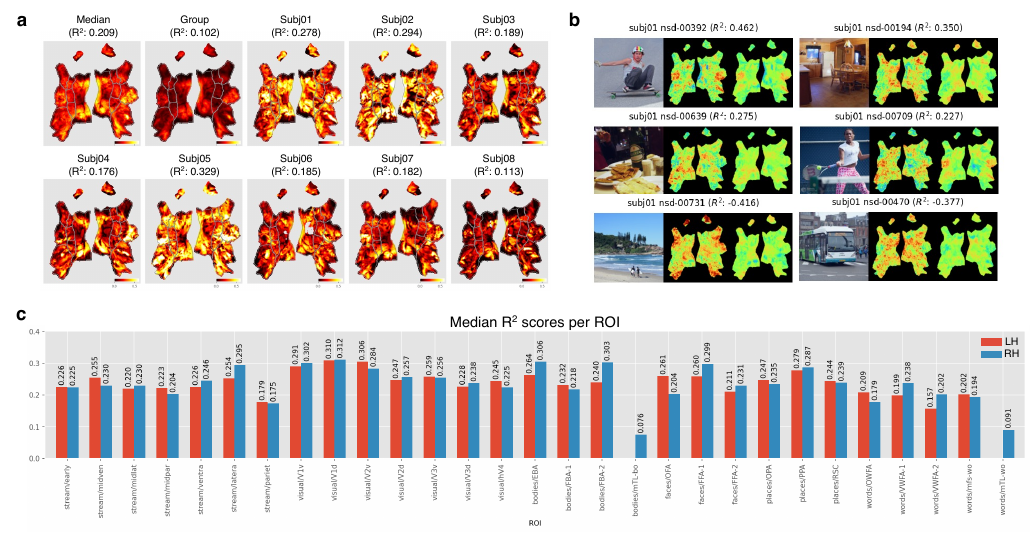}
\caption{\textbf{Prediction performance.} \textbf{a} Vertex-wise $R^2$ maps for each subject on the validation split. \textbf{b} Individual sample images (left), fMRI targets (middle), and predictions (right) for NSD Subject 1. \textbf{c} Group median $R^2$ scores for individual ROIs.}
\label{fig:pred_perf}
\end{figure*}

\paragraph{Subject prediction maps.}
In Figure~\ref{fig:pred_perf} \textbf{a}, we visualize each subject's reconstruction $R^2$ values, projected onto a cortical flat map \cite{gao2015pycortex}. We observe substantial variation in prediction performance across subjects, which appears correlated with the number of sessions each subject completed \cite{Allen2022}.

The spatial pattern is largely consistent across subjects. This is reflected in the group-wise median map (top left). Early visual cortex as well as the face, body, and place selective regions of the ventral stream are all consistently well predicted. In each subject, we observe worse prediction in regions susceptible to fMRI signal loss, as well as along the apparent boundaries between functionally well-defined areas.

Finally, we note that the ``Group'' $R^2$ map, which reflects subject \textit{agnostic} predictions (i.e.\ excluding the subject-specific path in Figure~\ref{fig:arch}), captures significant variation in the data. The spatial profile is consistent with the individual subject maps, albeit dampened.

\paragraph{Individual sample predictions.}
In Figure~\ref{fig:pred_perf} \textbf{b}, we visualize predictions for individual samples in a single well-predicted subject (Subject 1). The selected examples roughly span the range of observed reconstruction scores. The most well-predicted examples are often clear and unambiguous images of people playing sports or indoor and outdoor scenes. But there are also several samples for which the predicted activations are highly \textit{anti}-correlated (bottom row). These appear to be samples with large signal variation. The presence of negatively predicted examples is somewhat surprising, and may reflect the model's high degree of regularization. Alternatively, there could be unaccounted sources of signal \cite{yang2023memory}.

\paragraph{ROI prediction performance.}
In Figure~\ref{fig:pred_perf} \textbf{c}, we show group median $R^2$ scores for each ROI included in the challenge parcellation. At the coarsest granularity, the lateral stream ROI is predicted the best, while the parietal stream is predicted worst. At a finer scale, primary visual cortex (bilateral V1v/d) as well as right lateralized body and face selective regions are predicted best, whereas mid-level visual areas (V3v/d, V4) and word-selective regions are predicted worse.

\section{Conclusion}

In this report, we described our submission to the Algonauts 2023 challenge (``BlobGPT'') and presented some experimental results. Overall, our submission represents a strong baseline with a relatively simple and parameter-efficient architecture and minimal training pipeline. The main idea underlying our model was to conserve parameters by sharing wherever possible. The main takeaways from our experiments were that (1) end-to-end fine-tuning appears only modestly useful, (2) a rich supervision signal may be similarly important as image diet for predicting brain activity, and (3) visual cortex prediction performance follows the well-defined regional boundaries.

\paragraph{Acknowledgments.} We thank Sam Alldritt, Nathalia Bianchini Esper, Teresa George, Steven Giavasis, Amy Gutierrez, Helya Honarpisheh, Elizabeth Kenneally, Bene Ramirez, Florian Rupprecht, Zexi Wang and Ting Xu for helpful discussions on this work. We would also like to thank Paul Scotti and the MedARC fMRI project team for helpful discussion on the challenge in general. We thank the Pittsburgh Supercomputing Center (PSC) for granting GPU compute hours on the Bridges2 cluster.

{\small
\bibliographystyle{ieee_fullname}
\bibliography{report}
}

\end{document}